%% file: Paper.tex
\documentclass[10pt,twocolumn,letterpaper]{article}

\usepackage{cvpr}              
\usepackage{graphicx}
\usepackage{amsmath}
\usepackage{amssymb}
\usepackage{booktabs}
\usepackage{textcomp}

\usepackage[pagebackref,breaklinks,colorlinks]{hyperref}
\usepackage[font=small, skip=3pt]{caption}

\usepackage[capitalize]{cleveref}
\crefname{section}{Sec.}{Secs.}
\Crefname{section}{Section}{Sections}
\Crefname{table}{Table}{Tables}
\crefname{table}{Tab.}{Tabs.}

\def\cvprPaperID{4882} 
\def\confName{CVPR}
\def\confYear{2022}

\input{preamble}
\begin{document}

\title{ActiveZero: Mixed Domain Learning for Active Stereovision with Zero Annotation}
\author{Isabella Liu$^1$ \hspace{0.5cm} Edward Yang$^1$ \hspace{0.5cm} Jianyu Tao$^{1}$ \hspace{0.5cm} Rui Chen$^2$ \hspace{0.5cm} Xiaoshuai Zhang$^1$\\
Qing Ran$^3$ \hspace{0.5cm} Zhu Liu$^3$ \hspace{0.5cm} Hao Su$^1$\\
\\
$^1$University of California, San Diego \hspace{0.5cm}  $^2$Tsinghua University \hspace{0.5cm} $^3$Alibaba DAMO Academy
}
\maketitle
\begin{abstract}
Traditional depth sensors generate accurate real world depth estimates that surpass even the most advanced learning approaches trained only on simulation domains. Since ground truth depth is readily available in the simulation domain but quite difficult to obtain in the real domain, we propose a method that leverages the best of both worlds. In this paper we present a new framework, ActiveZero, which is a mixed domain learning solution for active stereovision systems that requires no real world depth annotation. First, we demonstrate the transferability of our method to out-of-distribution real data by using a mixed domain learning strategy. In the simulation domain, we use a combination of supervised disparity loss and self-supervised losses on a shape primitives dataset. By contrast, in the real domain, we only use self-supervised losses on a dataset that is out-of-distribution from either training simulation data or test real data. Second, our method introduces a novel self-supervised loss called temporal IR reprojection to increase the robustness and accuracy of our reprojections in hard-to-perceive regions. Finally, we show how the method can be trained end-to-end and that each module is important for attaining the end result. Extensive qualitative and quantitative evaluations on real data demonstrate state of the art results that can even beat a commercial depth sensor.
\end{abstract}

\section{Introduction}
\input{sec/Intro}

\section{Related Work}
\input{sec/RelatedWork}

\section{Method}
\input{sec/Method}

\section{Experiments}
\input{sec/Experiment}

\section{Conclusion and Future Work}
\input{sec/Conclusion}

{\small
\bibliographystyle{ieee_fullname}
\bibliography{citation}
}

\end{document}


\title{Supplementary Material for ``ActiveZero: Mixed Domain Learning for Active Stereovision with Zero Annotation''}

\maketitle

\section {Additional Ablation Study}

\subsection{Effect of Simulation Ground-truth}

\begin{figure}[h]
\begin{center}
	\includegraphics*[width=3in]{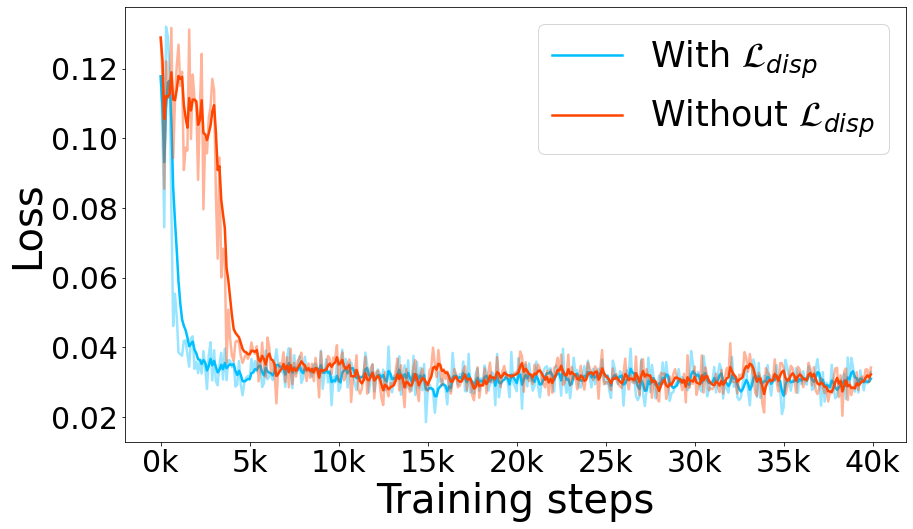}
\end{center}    
  \caption{Loss curve of training with and without simulation disparity ground-truth}
\label{fig:supp_a_1}
\end{figure}

In this section, we study the effect of using the supervised simulation disparity loss $\mathcal{L}_{disp}$ during training. To do so, we conduct experiments with and without $\mathcal{L}_{disp}$ added to the final loss term and observe their convergence rate as well as final converged solution. \Cref{fig:supp_a_1} shows that adding simulation disparity loss (blue) helps the network converge faster to the global optima.

\subsection{Patch Size of Reprojection Loss}

\begin{figure}[h]
\begin{center}
	\includegraphics*[width=3in]{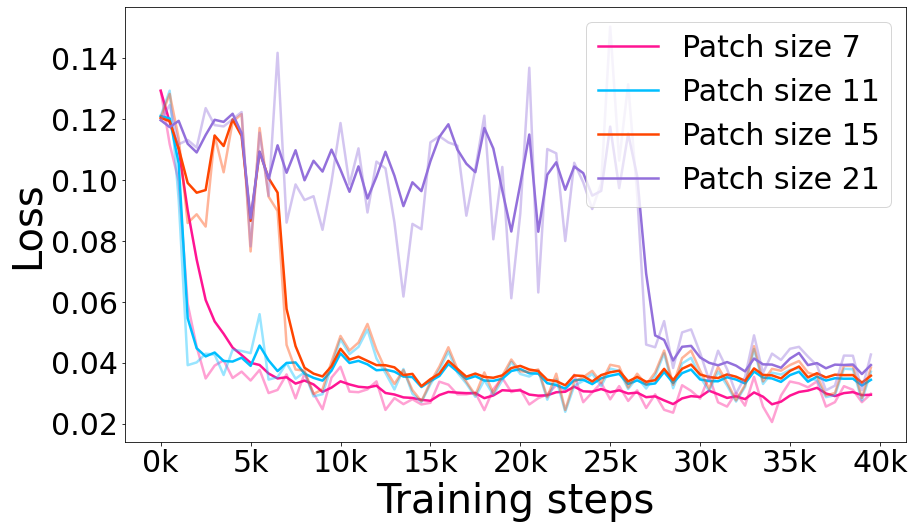}
\end{center}    
  \caption{Loss curve of training using different patch sizes}
\label{fig:supp_a_2}
\end{figure}

In this section, we conduct an ablation study on the patch size of the patch-wise reprojection loss. In the main paper, we chose a patch size of $11$. For this study, we change patch size to $7$, $15$ and $21$, train each one with only the real reprojection loss term, and evaluate them on the same testing dataset. \Cref{tab:supp_a_2} suggests patch size $15$ has the best result on the absolute depth error (\textit{abs depth err}) metric while patch size $21$ has the lowest percentage of depth outliers with absolute depth error larger than 4mm (\textit{$>$4mm}). However, the loss curve in \cref{fig:supp_a_2} indicates that patch size $11$ converges faster than the other patch sizes. Considering patch size $11$ also occupies less GPU memory during training, we choose patch size $11$ in our main experiments. 

\begin{table}[h]
\centering
\begin{tabular}{c|c|c}
\toprule
Patch size & Abs depth err (mm) $\downarrow$ & $>$ 4mm $\downarrow$\\ \hline 
7 & 5.507 & 0.466           \\
11 & 5.115 & 0.393          \\
15 & \textbf{5.114} & 0.386 \\
21 & 5.402 & \textbf{0.385} \\
\bottomrule
\end{tabular}
\caption{Performance of different patch size}
\label{tab:supp_a_2}
\end{table}

\subsection{Loss Ratio between Simulation and Real Domain}

In this section, we conduct an ablation study on the loss weight $\lambda_s$ and $\lambda_r$ described in Sec.~3.3 of the main paper. In our main experiment, we use $\lambda_s = 0.01$ and $\lambda_r = 2$. We change $\lambda_s$ and $\lambda_t$ to different values and test the trained models on the testing dataset. The results in \cref{tab:supp_a_3} indicate that when $\lambda_s = 0.01$ and $\lambda_r = 2$, the network achieves the best result, which is consistent with our experiment setting.

\begin{table}[h]
\centering
\begin{tabular}{c|c|c|c}
\toprule
$\lambda_s$ & $\lambda_r$ & Abs depth err (mm) $\downarrow$ & $>$ 4mm $\downarrow$\\ \hline 
1 & 0.5 & 7.578 & 0.548     \\
1 & 1 & 6.064 & 0.455       \\
1 & 2 & 5.672 & 0.446       \\
0.05 & 2 & 5.543 & 0.433    \\
0.01 & 2 & \textbf{4.377} & \textbf{0.335}    \\
0.002 & 2 & 4.683 & 0.368   \\
\bottomrule
\end{tabular}
\caption{Performance of different loss weight}
\label{tab:supp_a_3}
\end{table}

\subsection{The 6-layer CNNs}

\begin{figure}[h]
    \centering
	\includegraphics[width=0.48\textwidth]{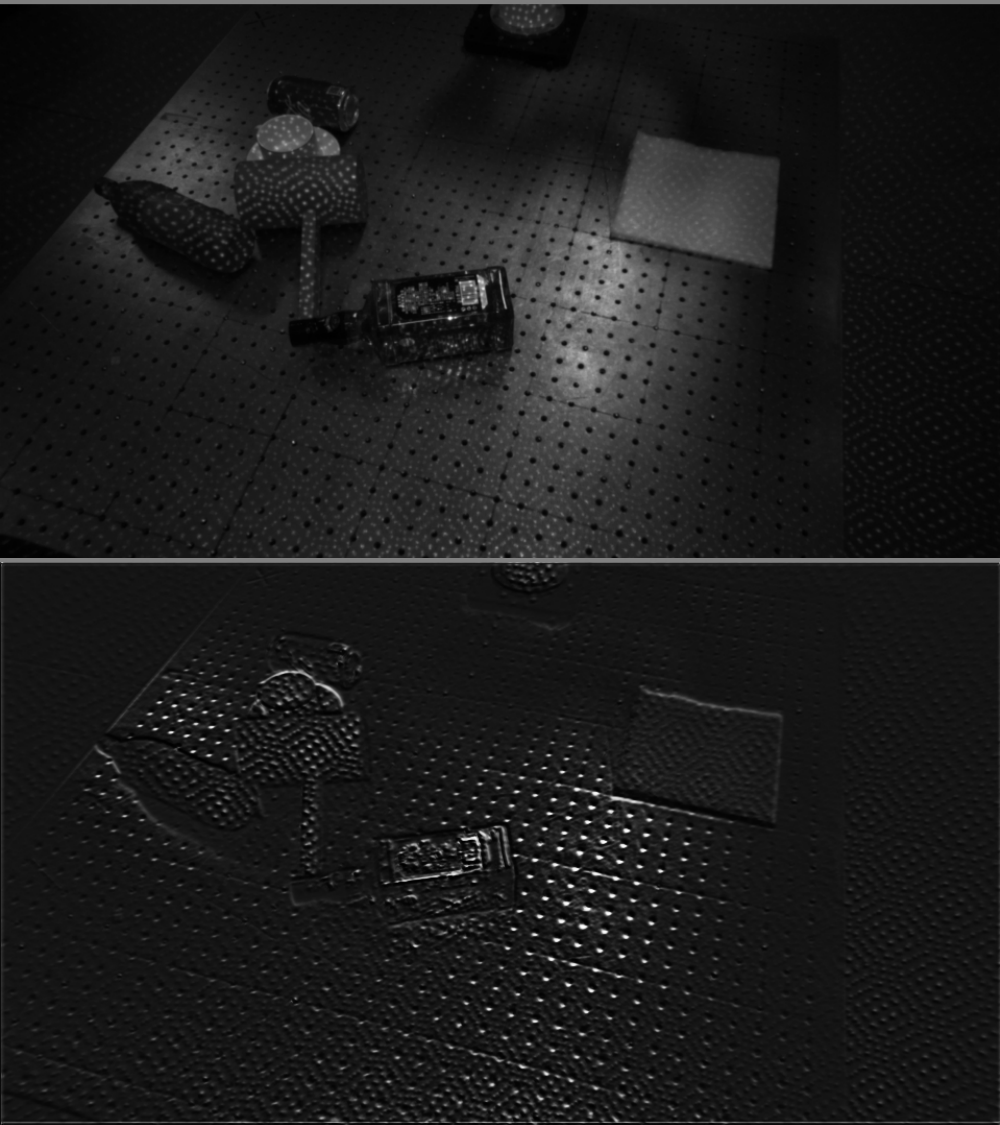}
	\caption{The effect of the 6-layer CNN filter. The top image is the captured IR image; the bottom image is the output of the 6-layer CNN filter. As shown, the lighting effect and the texture of the objects are reduced after passing through the filter.}
	\label{fig:filter}
\end{figure}

We experiment on the effectiveness of the 6-layer filter module in our proposed pipeline. As shown in \cref{tab:denoiser}, when training with the 6-layer filter, we achieve better performance than the pipeline without this module. The reason behind this is that this filter alleviates the lighting effect of the original image, in \cref{fig:filter}, so that the gap between the simulation dataset and real dataset decreases. 

\begin{table}[h]
\centering
\begin{tabular}{c|c|c}
\toprule
Method &  Abs depth err (mm) $\downarrow$ & $>$ 4mm $\downarrow$\\ \hline 
w/o 6-layer filter & 4.592 & 0.356 \\
6-layer filter & \textbf{4.377} & \textbf{0.335} \\
\bottomrule
\end{tabular}
\caption{Performance of network trained with 6-layer filter and without 6-layer filter.}
\label{tab:denoiser}
\end{table}

\subsection{Inference Time}

We measure the inference time of our proposed pipeline in \cref{tab:infertime}. Our method has an average inference time of 0.25 seconds per image pair with a resolution of 960$\times$540. Compared to StereoGAN with PSMNet backbone, our method achieves faster inference times while also having better performance. We will continue to reduce our inference time in future studies.

\begin{table}[h]
\centering
\begin{tabular}{c|c}
\toprule
Method &  Inference Time(s) $\downarrow$\\ \hline 
StereoGAN+PSM & 0.303  \\
Our Method & 0.256 \\
\bottomrule
\end{tabular}
\caption{Inference time of StereoGAN+PSM and our method}
\label{tab:infertime}
\end{table}

\section{More Details of Datasets}
\begin{figure*}[h]
\hspace{.23\linewidth}\small{\textbf{RGB}}\hspace{.22\linewidth}\small{\textbf{IR}}\hspace{.21\linewidth}\small{\textbf{Disparity}}\\
\begin{subfigure}{\textwidth}
    \centering
	\includegraphics[width=0.75\textwidth]{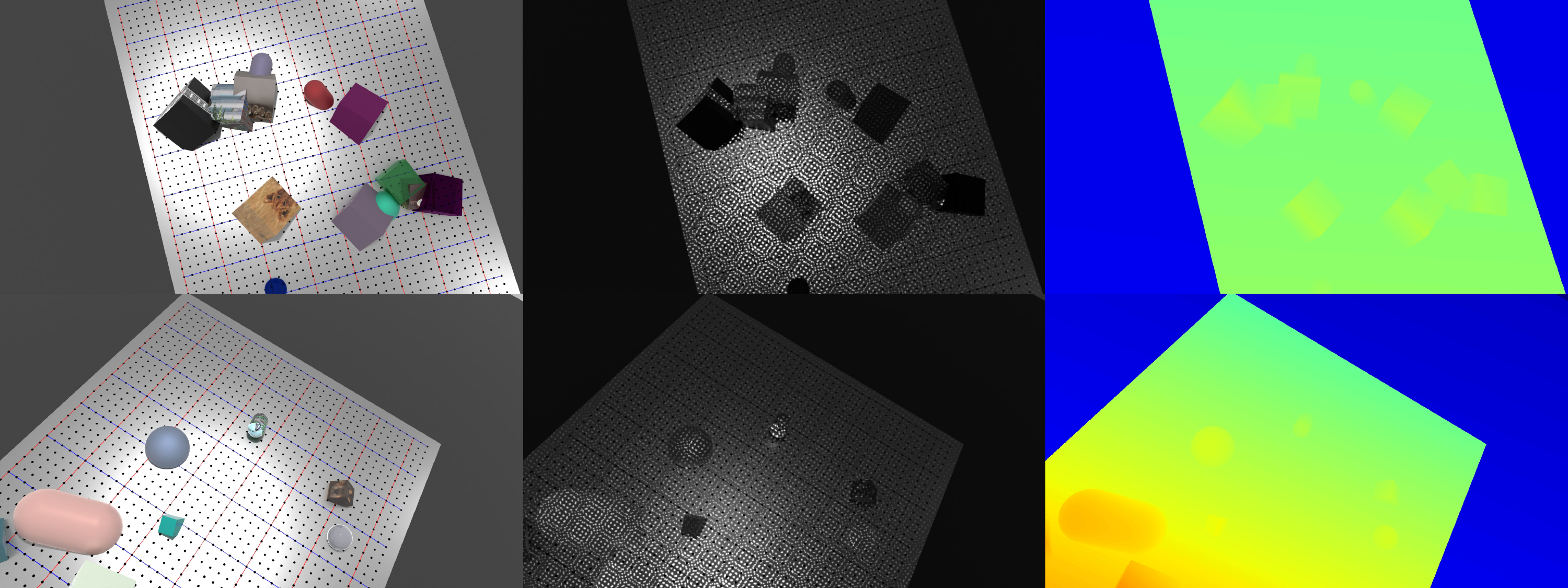}
	\caption{}
\end{subfigure}
\begin{subfigure}{\textwidth}
    \centering
	\includegraphics[width=0.75\textwidth]{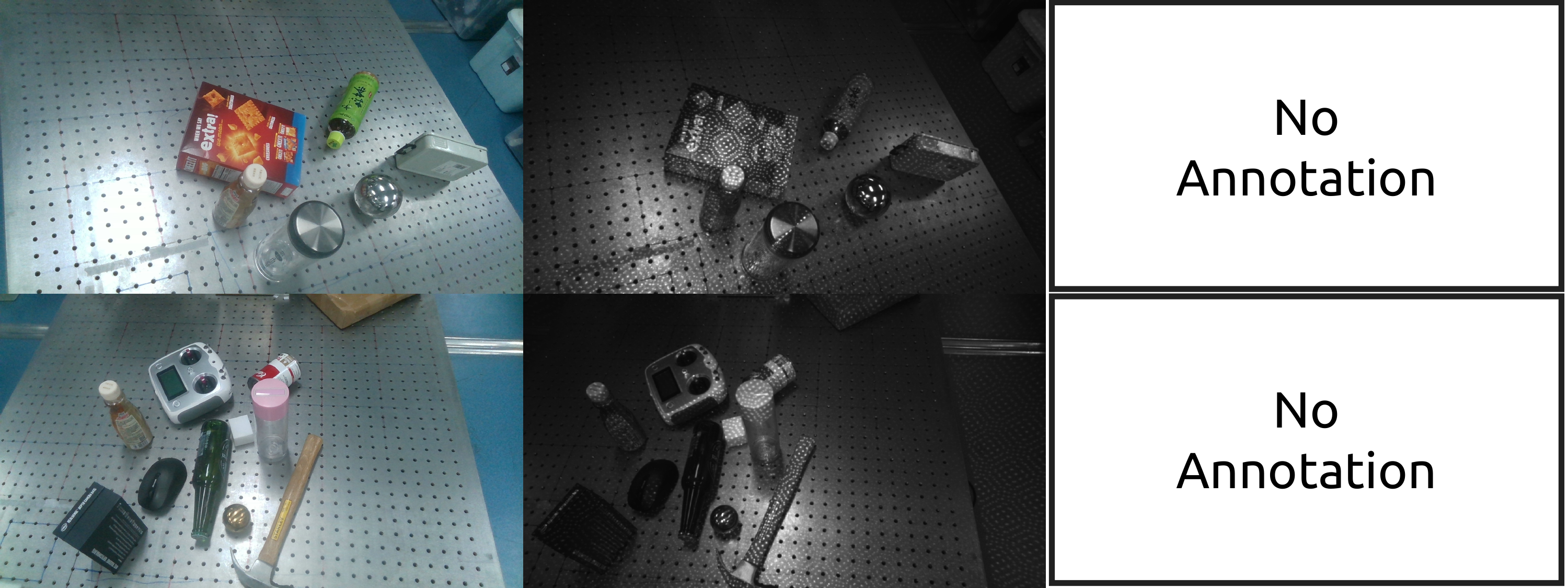}
	\caption{}
\end{subfigure}    
\begin{subfigure}{\textwidth}
    \centering
	\includegraphics[width=0.75\textwidth]{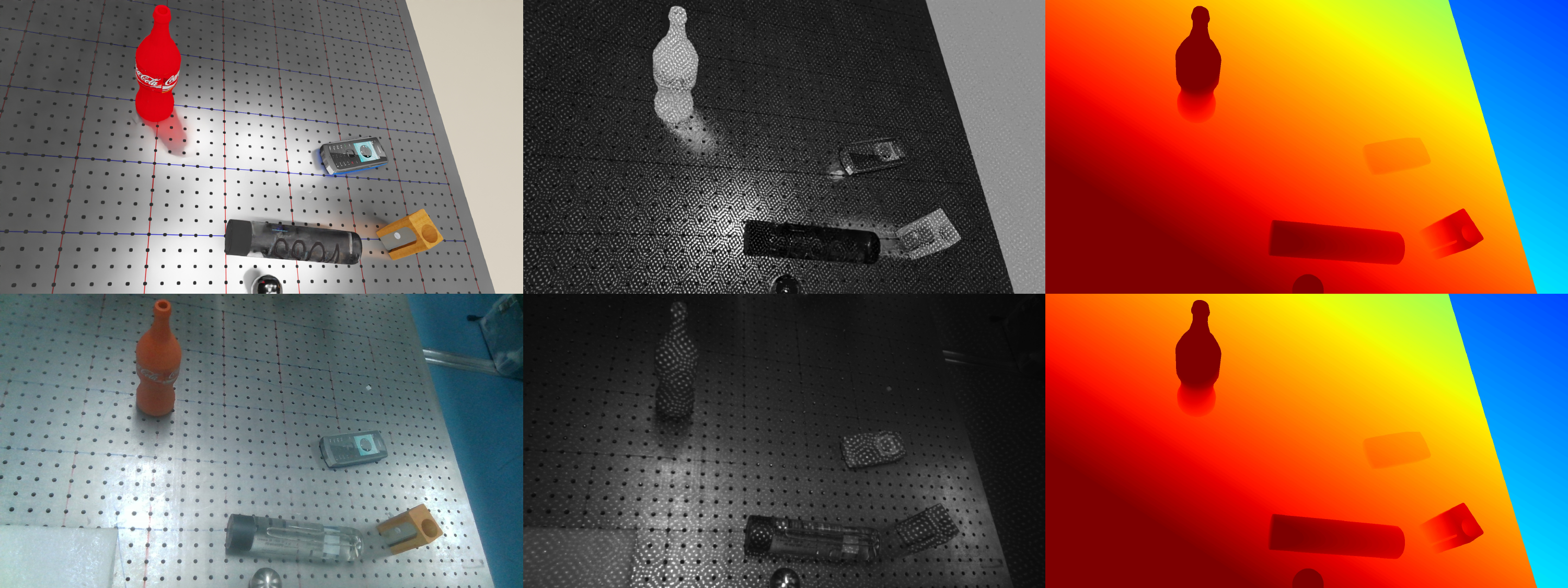}
	\caption{}
\end{subfigure}
\begin{subfigure}{\textwidth}
    \centering
	\includegraphics[width=0.75\textwidth]{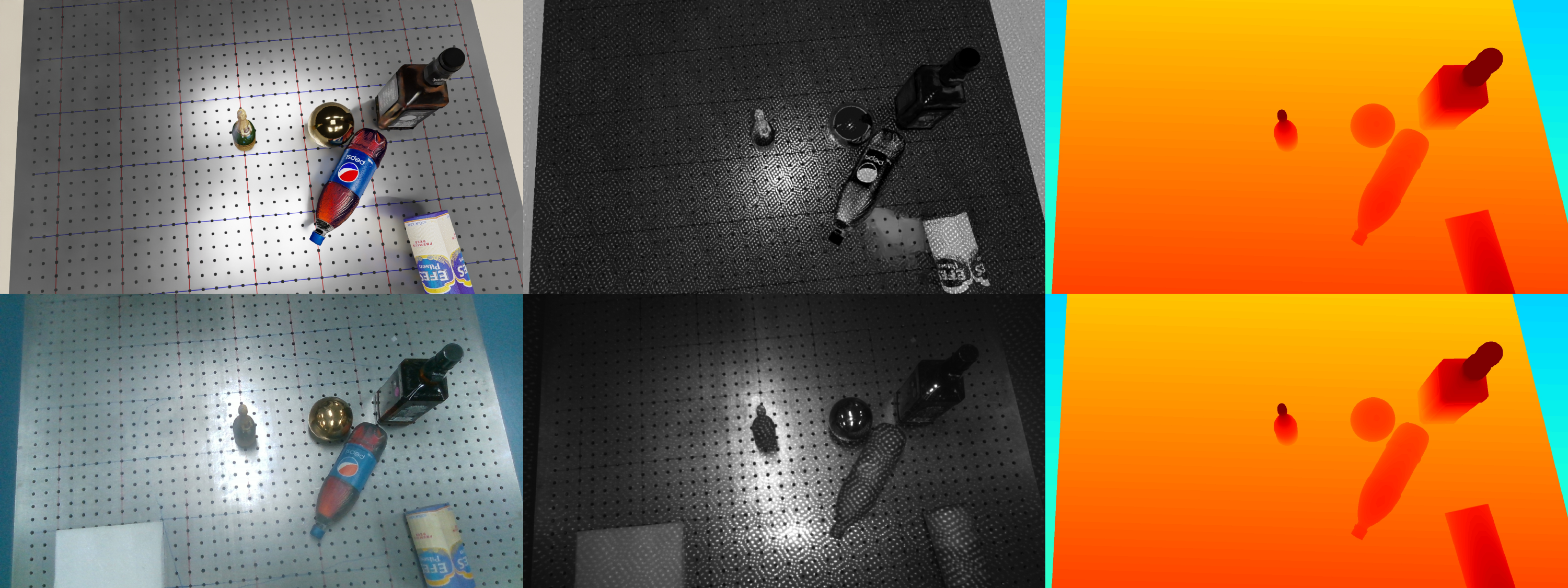}
	\caption{}
\end{subfigure}   
  \caption{More examples of our datasets. (a) is training simulation dataset, (b) is training real dataset, (c) and (d) are testing simulation and real pixel-wise aligned pairs.}
\label{fig:supp_dataset}
\end{figure*}

The training simulation dataset has 18000 image pairs with random camera extrinsics, shape primitives, textures and poses. As in \cref{fig:supp_dataset} (a), in order to make the scene more complicated, the primitives can overlap with each other and are not strictly attached to the table. Therefore, they can either overlap with the table or float above the table. In \cref{fig:supp_dataset} (a), the textures are randomly selected to improve generalizability. For IR images in \cref{fig:supp_dataset} (a), the simulated IR pattern is projected onto each scene of the simulation dataset.

Samples of the training real dataset are shown in \cref{fig:supp_dataset} (b). The objects in the training dataset are not present in the testing dataset and the ground truth depths are not required for this dataset. To preserve its generalizability, the optical properties of the objects are diversely selected. In \cref{fig:supp_dataset} (b), there exists objects that are transparent (glass bottle), specular (the cover of the glass bottle) and diffused (black paper box). These objects have different abilities to reflect IR pattern as seen in \cref{fig:supp_dataset} (b). Temporal IR images are collected by adjusting the power of the pattern emitter. There are 6 images with increasing IR power in each scene.

The testing dataset contains objects that are never used in training to best represent the generalizability of our method. As shown in \cref{fig:supp_dataset} (c) and (d), the object properties are also diversely selected. For example, this dataset contains specular objects (metal ball), transparent objects (bottled water) and diffused objects (printed cell phone). The IR pattern is collected by adjusting the IR emitter to the max power used in the training dataset. To obtain accurate ground truth, we align the scene using the same object poses and camera parameters in simulation, as shown in \cref{fig:supp_dataset} (c) and (d).

%% file: preamble.tex
\usepackage{overpic}
\usepackage{enumitem} 
\usepackage{overpic} 
\usepackage{color}
\usepackage{multirow}

\definecolor{turquoise}{cmyk}{0.65,0,0.1,0.3}
\definecolor{purple}{rgb}{0.65,0,0.65}
\definecolor{dark_green}{rgb}{0, 0.5, 0}
\definecolor{orange}{rgb}{0.8, 0.6, 0.2}
\definecolor{red}{rgb}{0.8, 0.2, 0.2}
\definecolor{darkred}{rgb}{0.6, 0.1, 0.05}
\definecolor{blueish}{rgb}{0.0, 0.3, .6}
\definecolor{light_gray}{rgb}{0.7, 0.7, .7}
\definecolor{pink}{rgb}{1, 0, 1}
\definecolor{greyblue}{rgb}{0.25, 0.25, 1}

\usepackage{blindtext}

\renewcommand{\paragraph}[1]{\vspace{1em}\noindent\textbf{#1}.}

%% file: sec/Intro.tex
Depth sensors can provide 3D geometry information about a target scene, which is critical in various robotic applications, including mapping, navigation, and object manipulation~\cite{Mishra2014RobotAM, hwang2020applying, cunha2011using}. Among the different types of depth sensors available, active stereovision depth sensors (eg. Intel RealSense\texttrademark D series) are the most widely adopted in both industry and academic settings due to their high spatial resolution, high accuracy, and low cost~\cite{keselman2017intel}. These sensors are composed of an infrared (IR) pattern emitter and two IR cameras with the IR pattern projected onto the target scene to facilitate stereo matching. However, since these sensors use classical stereo algorithms, they suffer from common stereo matching issues such as over smoothing, edge fattening and holes for specular and transparent objects so they are non-ideal for robotic applications which require high precision and completeness~\cite{CHEN2022106763}. 

\begin{figure}[t]
\begin{center}
	\includegraphics*[width=.48\textwidth]{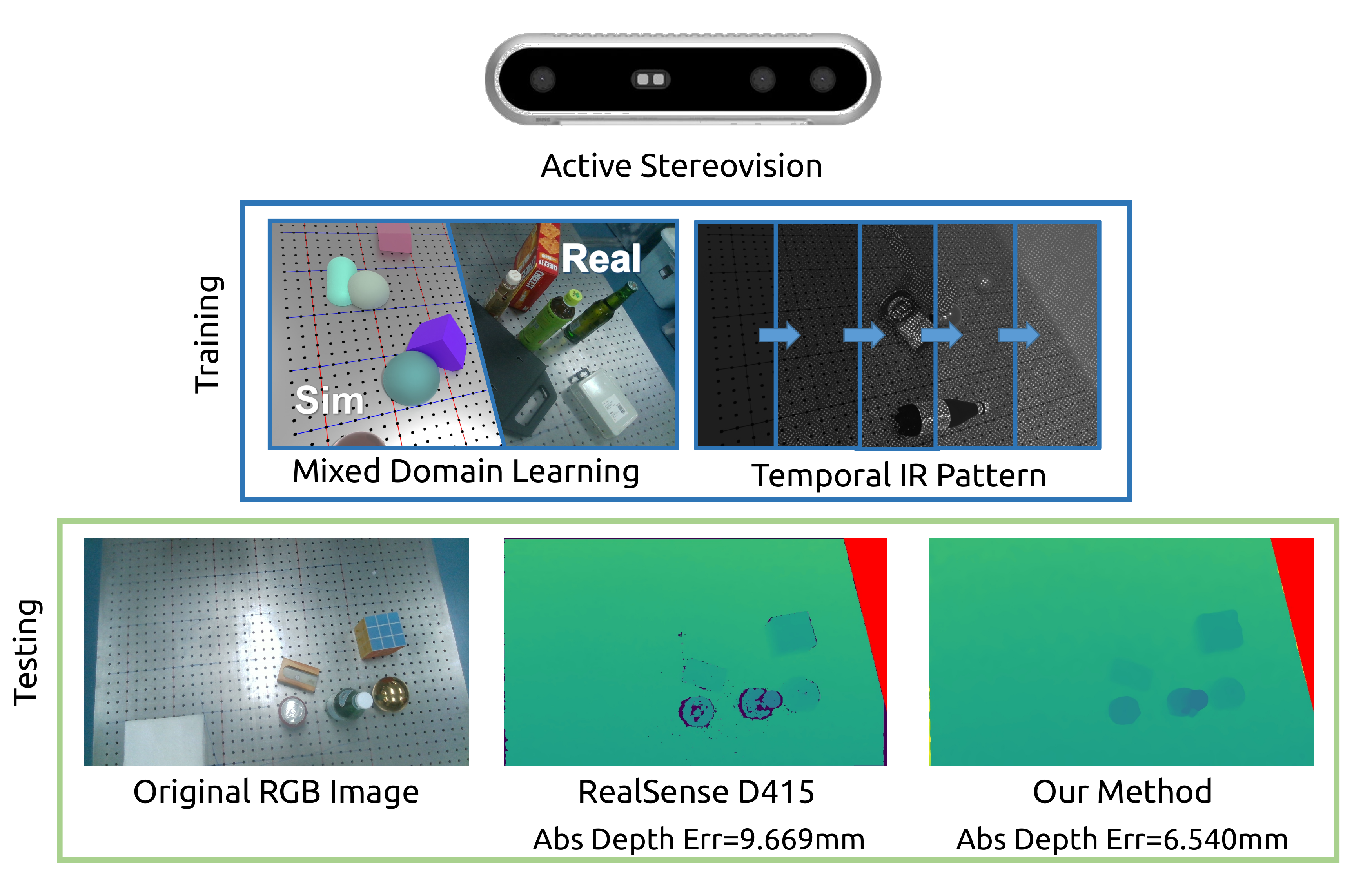}
\end{center}    
  \caption{ActiveZero produces more accurate and complete disparity estimates on real IR stereo images for objects with complex optical characteristics (specular, transparent) than commercial depth sensors with zero real depth annotation using \textit{mixed domain learning} by leveraging self-supervised reprojection loss on temporal IR patterns in the real domain and direct disparity supervision in the simulation domain.}
\label{fig:introduction}
\end{figure}

Learning based methods can solve the aforementioned issues by generating more accurate and complete depth maps through the utilization of prior samples to understand how to correctly handle edges and uncertain pixels~\cite{chang2018pyramid, yao2018mvsnet, chen2019point, chen2020visibility}. However, a large scale stereo dataset with ground truth depth is required to train these learning based methods, which is costly and time-consuming to collect in the real world. Therefore, one way to alleviate this problem is to use self-supervised learning. Self-supervised stereo methods ~\cite{zhang2018activestereonet, zhong2017selfsupervised} use reprojection or other related losses between binocular images as supervision, but the fluctuation of these losses prohibit the network from reaching a meaningful optima. Another approach is to use simulation data where ground truth depth is readily available. However due to the domain gap between the simulation and real world, networks trained on only simulation data cannot be reliably transferred to the real domain. Domain adaptation methods have been proposed to overcome the Sim2Real problem~\cite{liu2020stereogan}, but the introduction of GANs makes the training process unstable~\cite{kodali2017convergence} and the performance suboptimal. 

This paper proposes an end-to-end learning stereo method that combines the advantages of self-supervised learning in the real domain and supervised learning in the simulation domain which we call \textit{mixed domain learning} (\cref{fig:introduction}). This strategy significantly boosts the stereo network performance while also stabilizing and speeding up the optimization process. Specifically, by only needing to train on shape primitives in the simulation domain with ground truth depth as supervision and an unrelated set of scenes in the real domain with reprojection as self-supervision, we are able to achieve comparable performance on completely out-of-distribution objects in the real domain as though we were directly training on those objects. 

\begin{figure*}[t]
\begin{center}
	\includegraphics*[width=0.95\textwidth]{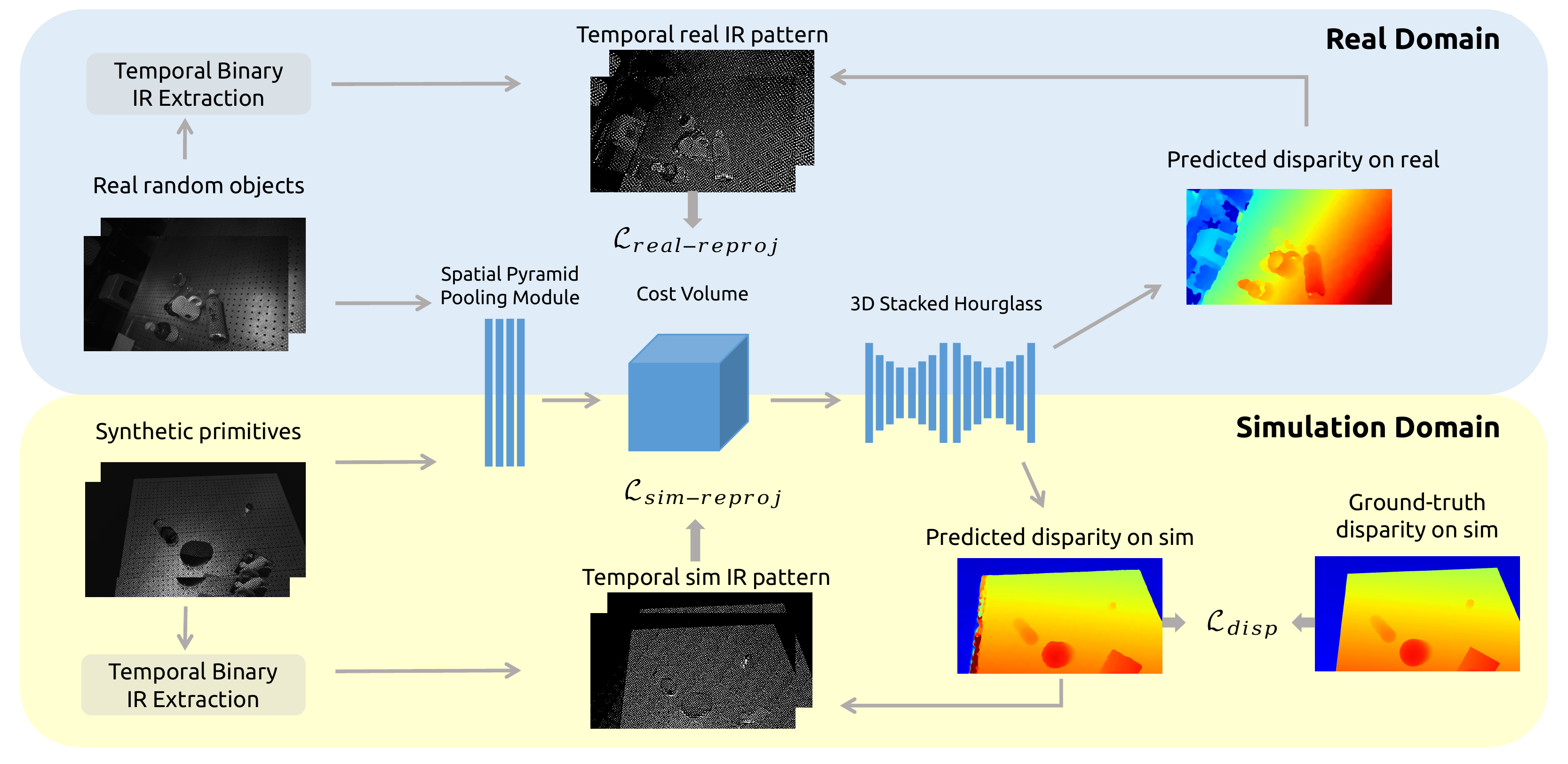}
\end{center}    

  \caption{Architecture overview. The simulated and real stereo IR images are fed to a shared weight stereo network consisting of a CNN for noise reduction and a cost-volume-based 3D CNN for disparity prediction. The network is trained with reprojection loss on temporal binary IR pattern in the real domain,  reprojection loss and disparity loss in the simulation domain as mixed domain learning.}
\label{fig:architecture}
\end{figure*}

In addition, we observed that there are fundamental issues with performing direct image reprojection as previous works had done so we propose the use of temporal IR by periodically adjusting the brightness of the emitted IR pattern and extracting the binary pattern from the temporal image sequences. The reprojection loss on the temporal binary pattern eliminates the influence of scene texture and also the effect of illumination strength decaying with increased distance. Experimental results demonstrate that our method is able to outperform state-of-the-art learning-based stereo methods and commercial depth sensors, and ablation studies verify the effectiveness of each module in our work.

%% file: sec/RelatedWork.tex
\noindent\textbf{Depth sensors} can be classified into four categories according to their underlying sensing principle~\cite{CHEN2022106763}: passive stereo-vision, active stereo-vision, structured light, and time-of-flight. Each depth sensing technique has its own advantages and drawbacks. Giancola \textit{et al.}~\cite{giancola2018survey} introduces the principles of different depth sensors and evaluated their metrological performance independently. Chen \textit{et al.}~\cite{CHEN2022106763} compared the short-range depth sensing performance of 8 commercially available depth sensors for different illumination settings and objects and found that active stereovision sensors and structured light sensors have similar performance to each other and better performance than the other two kinds of sensors. Furthermore, depth sensor performance varies among different objects with these sensors performing especially poorly on objects with complex optical characteristics~\cite{sajjan2020clear}. In this paper, we focus on improving the visual and numerical performance of active stereovision depth sensors, but the framework can also be applied for structured light sensors.

\noindent\textbf{Learning Based Stereo} has become much more prevalent with large-scale benchmarks and higher computational ability~\cite{kitti, NIPS2012_c399862d, he2015deep}. Stereo matching for depth estimation is typically done in four steps: matching cost computation, cost aggregation, optimization, and disparity refinement~\cite{10011856636}. Zbontar and LeCun were the first to design a network for computing matching costs by utilizing a deep Siamese architecture~\cite{2015lecun}. Building on this, DispNet introduced the first end-to-end framework for predicting entire disparity maps from stereo image pairs~\cite{mayer2016large}. Works such as GWCNet followed and improved on this framework by using 3D convolutions to compute better cost volumes~\cite{kendall2017end}. Recent works have improved performance even further by utilizing multi-scale context aggregation to estimate depth at different resolutions in order to leverage global image-level information~\cite{chang2018pyramid, gu2020cascade}. However, the requirement of ground truth depth as supervision has limited the application of learning based stereo.

\noindent\textbf{Self-supervised Stereo} is a popular approach for stereo matching when ground truth depth is unavailable. Godard \textit{et al.}~\cite{godard2017unsupervised} explored the use of left-right consistency in a rectified stereo image pair for self-supervision. They reconstruct the right view based on the given left view and its predicted disparity map and then use the reconstruction loss as a supervision for training. PDANet~\cite{app11125383} introduced the idea of perceptual consistency to improve reconstruction quality on regions with low texture and high color fluctuations. ActiveStereoNet~\cite{zhang2018activestereonet} used local-contrast-normalized (LCN) reprojection loss on IR images as self-supervision to train a stereo network. However, this reprojection loss fluctuates along the epipolar line and is heavily influenced by occlusion and viewpoint variance. Not only that, LCN loss also suffers in areas where camera noise and environmental illumination dominate the projected IR pattern since it only uses the IR image with projected pattern. Our method addresses these concerns using temporal IR reprojection loss by way of actively adjusting the brightness of the emitted IR pattern which is more robust to camera noise and environmental illumination. 

\noindent\textbf{Domain Adaptation} techniques have shown great promise in closing the gap between the simulation and real domains. Tobin \textit{et al.} ~\cite{tobin2017domain} proposed using domain randomization through randomizing rendering in the simulator to train a robust model that would interpret the real domain as just another variation of the simulation domain. Previous works have also tried aligning the source and target domains by matching their input distributions or their feature statistics~\cite{Long2013TransferSC, sun2016deep}. Other works have attempted to learn domain- invariant representations by augmenting the input based on certain criterion set forth in the task and approach itself~\cite{ganin2015unsupervised}. Moreover, unsupervised losses have seen increased use for domain adaptation in tasks such as semantic segmentation and object detection~\cite{tsai2020learning,sankaranarayanan2018learning,deng2018imageimage}.

Our work is most related to StereoGAN~\cite{liu2020stereogan}, which uses ground truth depth maps in the simulated domain and reprojection loss in the real domain along with unsupervised GAN losses in order to close the domain gap between simulation and real images. Our work differs from theirs in three key ways: (1) we utilize IR images with actively projected patterns for stereo matching instead of passive RGB images, which leads to a smaller sim2real gap and better transferability; (2) we use the proposed temporal  IR  reprojection  loss as self-supervision which is more effective in correlating local matching features; (3) we train using only shape primitives and random real objects that are out-of-distribution from test time data.

%% file: sec/Method.tex
In this section, we introduce \textit{mixed domain learning} for active stereovision. We first define the task setup: in real domain $\mathcal{X}$, we have a target set of real IR stereo images with projected pattern $\mathbb{X}^t = \{ (x^t_l, x^t_r)_i \}_{i=1}^N$, and our goal is to learn an accurate disparity estimation network $F$ to estimate the disparity $\hat x^t_d = F(x^t_l, x^t_r)$. We utilize \textit{mixed domain} data to train the network: in real domain $\mathcal{X}$ we collect another set of real IR stereo images $\mathbb{X} = \{ (x_l, x_r)_i \}_{i=1}^M$ without disparity annotation. To be clear, the objects appearing in $\mathbb{X}$ are different from the ones in $\mathbb{X}^t$. In simulation domain $\mathcal{Y}$, we generate a set of synthetic IR stereo images with ground truth disparity annotation $\mathbb{Y} = \{ (y_l, y_r, y_d)_i \}_{i=1}^K$. In order to guarantee the generalizability of the trained network to unseen objects,  we only use shape primitives (sphere, cube, capsule) with different scales, textures and materials to generate $\mathbb{Y}$.

\Cref{fig:architecture} shows the framework of our proposed method.
In the real domain, we propose the use of temporal binary IR reprojection loss as self-supervision (\cref{sec3.1}). In the simulation domain, we use the loss between predicted disparity and the ground truth disparity $y_d$ as supervision (\cref{sec:sec3.2}). The network is trained jointly using the self-supervision in real domain and supervision in simulation domain (\cref{sec:3.3}). The stereo network architecture and other implementation details are introduced in \cref{sec:3.4}.

\begin{figure}[t]
\begin{center}
	\includegraphics*[width=3.2in]{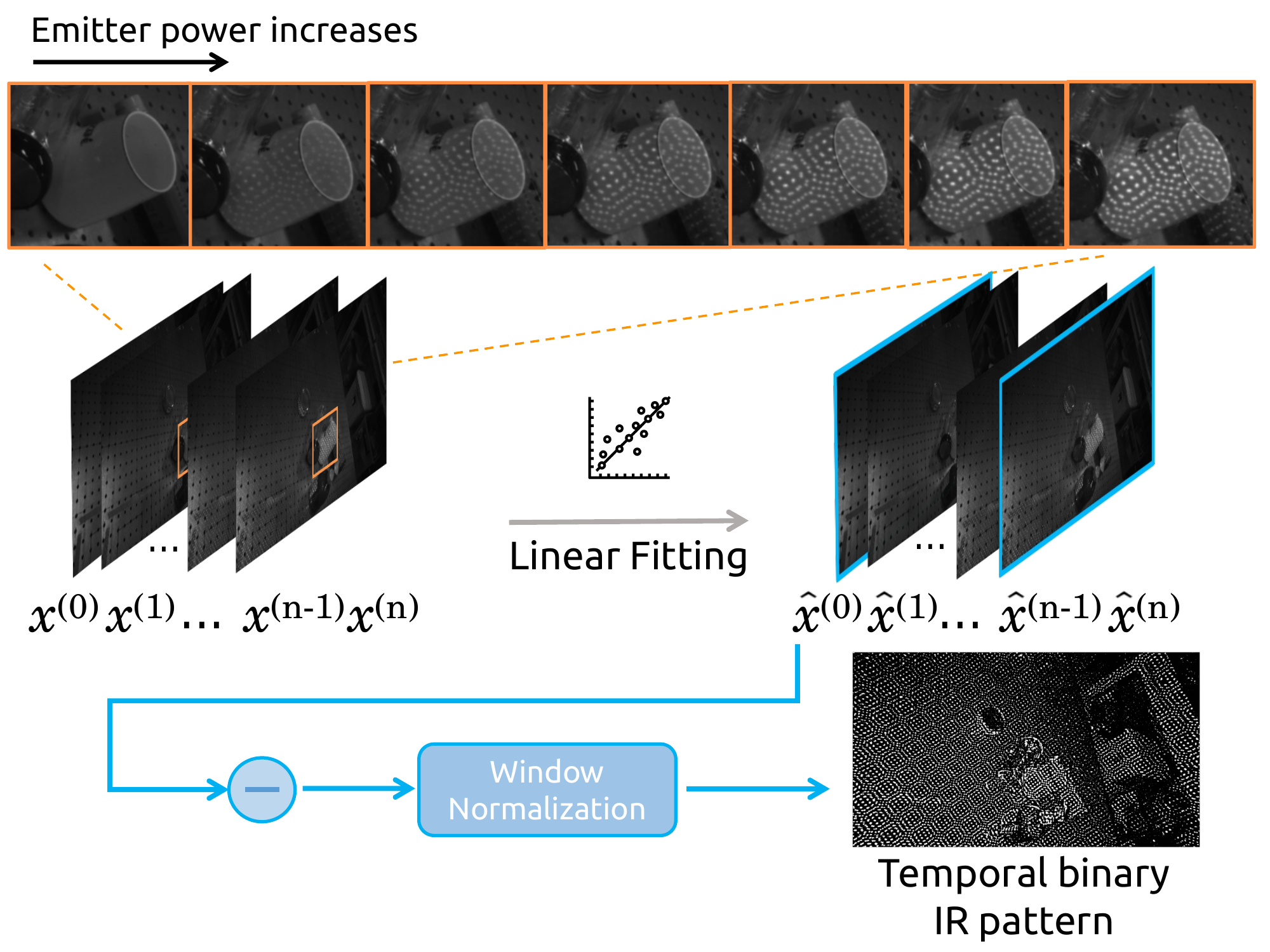}
\end{center}    
  \caption{Temporal binary pattern extraction}
\label{fig:extractor}
\end{figure}

\subsection{Real Domain: Self-supervised Learning on IR Images} 
\label{sec3.1}
The prerequisite for computing reprojection loss of grayscale stereo images in conventional self-supervised learning methods~\cite{godard2017unsupervised, zhang2018activestereonet} is that \emph{the object surface is Lambertian diffused} where the reflection intensity is invariant to the viewpoint, which is usually not satisfied in real world. Therefore, we propose to extract the binary projected active pattern from temporal IR stereo image sequences, which eliminates the adverse effect of surface reflection while maintaining the most important components of active pattern. Then, we construct the reprojection loss on this new binary pattern.

\begin{figure*}[t]
  \centering
  \includegraphics[width=1.0\linewidth]{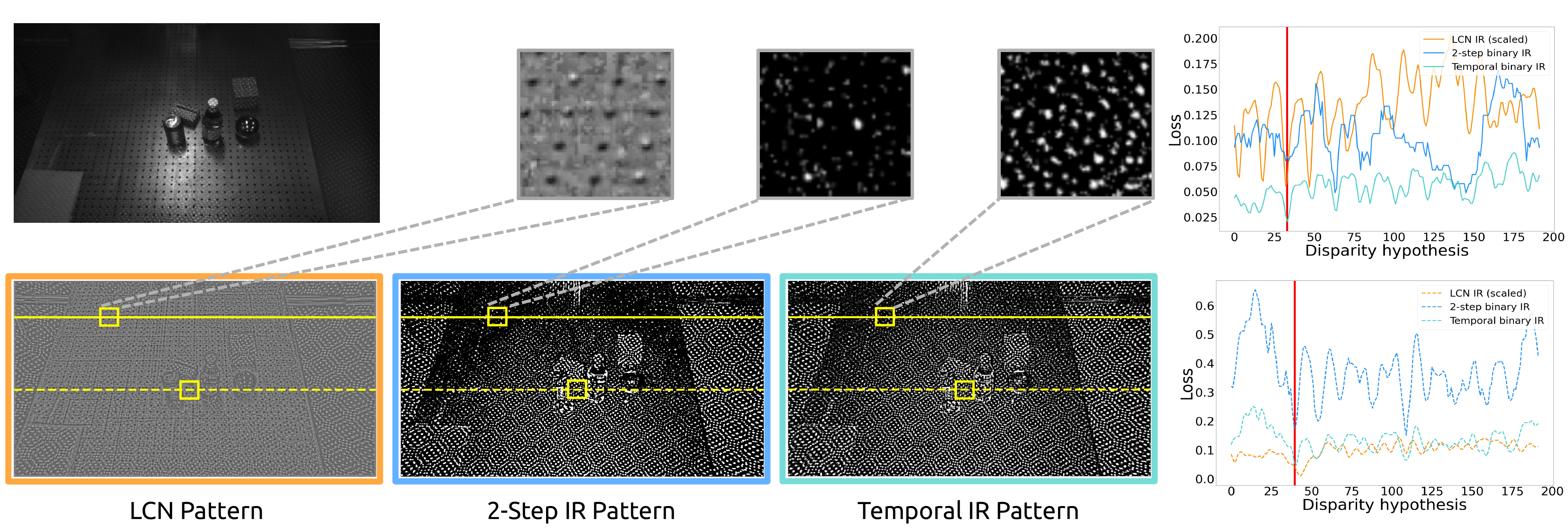}

\caption{Comparison of extracted pattern and reprojection loss along the epipolar line. LCN pattern represents local contrast normalization~\cite{zhang2018activestereonet} which consists of continuous values; 2-step IR pattern and temporal IR pattern represent the extracted binarized pattern from temporal IR image sequence using $n=1$ and $n=6$, respectively.}
   \label{fig:loss_curves}
\end{figure*}

\paragraph{Binary Pattern Extraction From Temporal IR Images}  
For the real captured IR images $x_l$ or $x_r$, the grayscale at pixel $(u,v)$ is:
\begin{equation}
    x_l(u,v) = I_l(u,v) + \alpha * e *K_l(u,v) + \epsilon
\end{equation}
where $I_l(u,v)$ represents the environmental illumination intensity, $K_l(u,v)$ represents the binary pattern captured by the camera, $\alpha$ represents the reflection coefficient determined by the object surface material, texture, angle and distance, $e$ represents the pattern emittance, and $\epsilon$ represents the camera sensor noise.
For active depth sensors, we manually adjust the pattern emittance $e$ by changing the emitter power. Therefore, as shown in \cref{fig:extractor}, our pattern extraction procedure is as follows:  we set $e$ to $\{e_0,e_1, ..., e_n\}$, capture a temporal sequence of corresponding IR images $\{x^{(0)}, x^{(1)}, ..., x^{(n)}\}$, and fit $x^{(0)}$,\ldots, $x^{(n)}$ to the linear model regressed and obtain $\hat x^{(0)}$, \ldots, $\hat x^{(n)}$. We extract the binary IR pattern $K(u,v)$ from the temporal image sequence through local window normalization and binarization:

\begin{equation*}
    K(u, v) = 
    \begin{cases}
        1 & ||\hat x^{(n)}(u, v) -  \hat x^{(0)}(u, v)|| > \delta(u, v) + c \\
        0 & otherwise \\
    \end{cases}
\end{equation*}

\begin{equation}
    \delta(u, v) = \frac{1}{w^2} \sum ||W(\hat x^{(n)}, u, v) - W(\hat x^{(0)}, u, v)||
\end{equation}
where $W(x, u, v)$ is a local window centered at pixel $(u,v)$ in $x$ with window size $w$, $c$ is a threshold to filter out noise and areas where the reflection coefficient is extremely small such as pure specular reflection regions. In our work, we use $n=6$.

In \cref{fig:loss_curves}, we compare the pattern extracted by different methods. By utilizing the temporal image sequence, our method is able to extract the pattern accurately and completely even in distant areas where the SNR (signal noise ratio) is low. The local normalization and binarization window filters out camera sensor noise and environmental illumination while retaining the projected active pattern, which is beneficial for further reprojection loss computation.

\paragraph{Binary Pattern Reprojection Loss} 
As demonstrated in traditional stereo matching and active stereo methods~\cite{mvs_tutorial, bleyer2011patchmatch, fanello2017ultrastereo, zhang2018activestereonet}, patch-wise reprojection losses are smoother and more accurate than pixel-wise losses and are beneficial for matching. Therefore, we construct the patch-wise reprojection loss on the extracted binary IR pattern $(K_l, K_r)$ :

\begin{equation*} 
\begin{aligned}
    &\mathcal{L}_{\text{reproj}} (K_l, K_r, \hat x_d) = \sum\limits_{uv}
    \frac{1}{{(2p+1)}^2} C(u,v) \\
    &C(u,v)=\sum\limits_{(u_p,v_p)  \in P(u,v)}||K_l(u_p,v_p)-\hat{K}_l(u_p,v_p)||^2\\
\end{aligned}
\label{real_reproj_loss}
\end{equation*}
where $P(u,v)$ represents the patch centered at pixel $(u,v)$ with patch size $(2p+1) \times (2p+1)$, $\hat{K}_l$ represents the warped right image using the predicted disparity $\hat x_d$. 

As shown in \cref{fig:loss_curves}, since the temporal binary IR pattern eliminates the influence of object texture and environmental illumination and only retains the projected pattern, the reprojection loss computed on the binary IR pattern reaches global minima at the ground truth disparity while the losses computed on the other two patterns could be misleading for the stereo network.

\subsection{Simulation Domain: Supervised Learning on Shape Primitives}
\label{sec:sec3.2}

Although the proposed temporal IR reprojection loss can be used
as the sole loss for stereo network training, it still has some limitations: the binary IR pattern cannot be extracted accurately for translucent and transparent objects and there are local minima in the loss with respect to the disparity hypotheses. On the other hand, traditional supervised learning with ground truth depth does not suffer from the aforementioned issues. However, it is costly and time-consuming to acquire ground truth depth in real world settings. Thus, we perform supervised learning only in the simulation domain. 

\paragraph{Dataset Generation based on Ray-tracing} In the last decade, there has been significant progress in ray-tracing rendering techniques in terms of speed and quality. Compared with rasterization, ray-tracing rendering can accurately simulate the light transmission process on translucent and transparent objects~\cite{pharr2016physically}.  Therefore, we use ray-tracing rendering to generate the simulated training dataset: we first build a cone lighting with mask to imitate the pattern emitter in the real active stereovision depth sensor, and then construct two cameras similar to stereo cameras in the real setting. The relative position between cameras and lighting are set using parameters from real sensors. We also add dim ambient light in the simulation environment to imitate the filtered environmental light in the real setting.

\paragraph{Shape Primitives} 
The semantic-specific biases in CAD model datasets may mitigate the generalizability of the learned stereo network. Thus we only use base shape primitives for simulated dataset generation. We use images from tiny ImageNet~\cite{le2015tiny} as object textures. The number of primitives is randomly sampled from 5 to 15. The sizes, layouts and materials are also randomly generated. 

\paragraph{Disparity loss} 
Given the synthetic stereo image pair with ground-truth disparity $(y_l,y_r,y_d)$, we follow \cite{chang2018pyramid} and adopt smooth $L_1$ loss between $y_d$ and the predicted disparity on synthetic stereo images:

\begin{equation}
    \mathcal{L}_{\text{disp}} = L_{1smooth}(F(y_l,y_r),y_d)
\end{equation}

\subsection{Mixed Domain Learning}
\label{sec:3.3}

Given the real stereo IR image$(x_l, x_r)$, and the simulated stereo IR image with ground truth disparity $(y_l,y_r,y_d)$, we train the stereo network $F(\cdot,\cdot)$ by combining the reprojection loss in the real domain and the disparity loss along with reprojection loss  in the simulation domain:
\begin{equation*}\label{loss}
\begin{aligned}
    \mathcal{L} (x_l, x_r, y_l, y_r, y_d) = & \lambda_r \cdot \mathcal{L}_{\text{real-reproj}} (x_l, x_r, F(x_l,x_r)) + \\
    & \lambda_s \cdot \Big [ \mathcal{L}_{disp} (F(y_l,y_r),y_d) +\\ & \mathcal{L}_{\text{sim-reproj}} (y_l, y_r, F(y_l,y_r)) \Big ] 
\end{aligned}
\end{equation*}
where $\lambda_r$ and $\lambda_s$ represent the weights of the real domain and the simulation domain respectively. 

The loss terms on real domain guarantee transferrability to unseen real data. However, we find that it is quite hard to train the network using these terms alone, due to noise in the self-supervision signals. Interestingly enough, after adding the supervised loss terms in simulation domain on primitive shapes, the behavior of loss minimization is much more tame: not only does the network converge faster, but also the final solution has better quality (see Sec.~A.1 in supplementary material and \cref{sec:exp:ablation} for empirical evidences).
\subsection{Implementation Details}
\label{sec:3.4}
In the stereo matching network, we adopt PSMNet~\cite{chang2018pyramid} as the backbone, which aggregates image features at different scales, constructs a cost volume and uses 3D CNNs to regress the disparity. The max disparity of PSMNet is set to be 192. We also use a 6-layer CNN to filter out irrelevant noise before feeding the stereo images into PSMNet. To make the model more robust, we apply color jitter and gaussian blur to the input images.

%% file: sec/Experiment.tex
\subsection{Experiment details}

\begin{figure}[t]
\hspace{.12\linewidth}\small{\textbf{RGB}}\hspace{.28\linewidth}\small{\textbf{IR}}\hspace{.22\linewidth}\small{\textbf{Disparity}}\\
\begin{subfigure}{.48\textwidth}
  \centering
  \includegraphics[width=\linewidth]{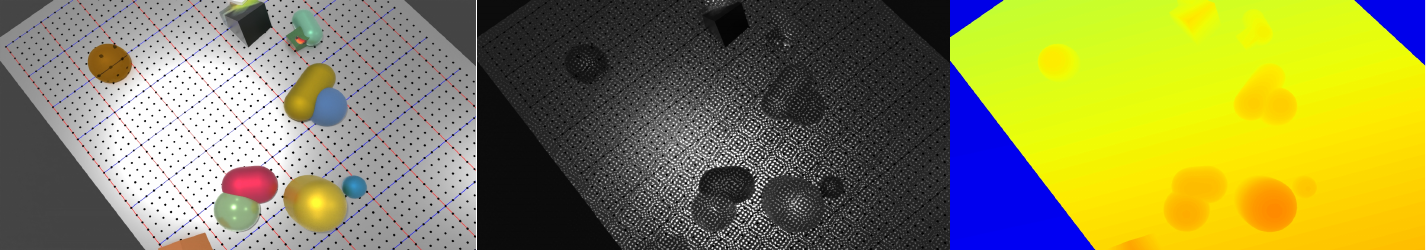}
  \caption{}
  \label{fig:sfig1}
\end{subfigure}
\begin{subfigure}{.48\textwidth}
  \centering
  \includegraphics[width=\linewidth]{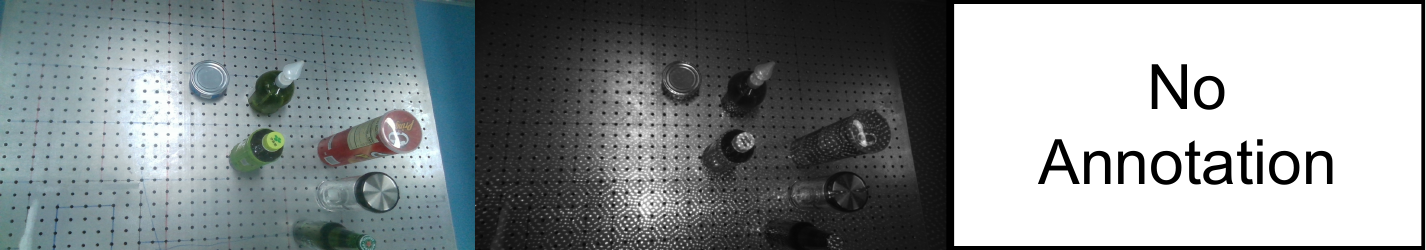}
  \caption{}
  \label{fig:sfig2}
\end{subfigure}
\begin{subfigure}{.48\textwidth}
  \centering
  \includegraphics[width=\linewidth]{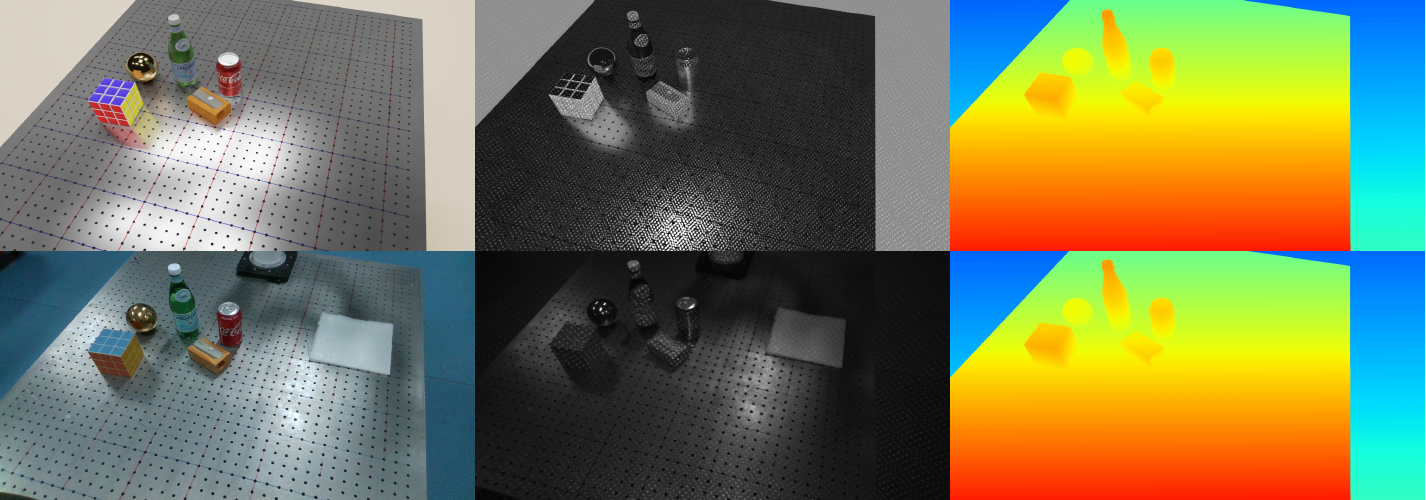}
  \caption{}
  \label{fig:sfig4}
\end{subfigure}

\caption{Example images from our dataset. (a) the simulation training dataset of random shape primitives; (b) the real training dataset of random objects different from testing; (c) the sim2real aligned testing dataset, including specular surfaces such as metals and translucent bodies such as liquids. Note: we don't rely on any annotation for real scenes which is why we have no disparity annotation in (b).}

\label{dataset}
\end{figure}

\paragraph{Datasets} 
\Cref{dataset} shows example images from the three datasets in our work. 
For the testing dataset, we used an Intel RealSense D415 as the active stereovision depth sensor. All the real RGB and IR images are captured using the RealSense camera. In order to quantitatively evaluate the performance of the camera, the complete and accurate ground depth is required. To do so,  we constructed a set of simulated scenes which are pixel-wise aligned with the real ones by precisely aligning the shapes and poses of objects and the intrinsic and extrinsic parameters of the RealSense camera. To evaluate the influence of object material on depth estimation performance, we include two categories of objects: 3D-printed objects and real objects. The 3D-printed objects are printed using color plaster powder, and are considered Lambertian diffused, while the real objects' material are complex (\textbf{specular, translucent, transparent}) and difficult for active stereovision depth sensors. Overall, the testing dataset consists of 504 stereo images of 24 different scenes. 

For the training dataset in the simulation domain, we rendered 20,000 stereo IR images with ground-truth disparity annotation using random shape primitives, including spheres, cubes and capsules. 10\% of the primitives are set to be transparent, 50\% are textured by images from tiny imagenet~\cite{le2015tiny}, and the rest are set to random colors. For the ray-tracing rendering, the number of samples per pixel is 128 and the max bounces is set to 8. The rendered IR images are post-processed by the NVIDIA OptiX denoiser~\cite{nvidiadeveloper_2021}.

For the training dataset in the real domain, we collected 1,047 real stereo IR images of random objects which are different from the testing dataset. The objects are randomly placed on the table, and captured by the same RealSense from different viewpoints. Note that we only use the real IR stereo images to construct the temporal IR reprojection loss, and the depth images are not collected.

\paragraph{Training} We train the network using the Adam optimizer with the initial learning rate set to 2e-4, decaying by half every 10k iterations for a total of 40k iterations. The network is trained on 2 GPUs each with 11GB GPU memory and a batch size of 4. We use $\lambda_s = 0.01$ and $\lambda_r = 2$ for the loss weight to set the two losses to similar scales.

For fair comparison, data augmentation is applied to both our method and baseline methods. Specifically, brightness and contrast is uniformly scaled by a value between 0.4 to 1.4, and 0.8 to 1.2 respectively. For gaussian blur, kernel size is fixed to $9\times9$ and the standard deviation is selected uniformly between 0.1 to 2. 

\paragraph{Evaluation metrics}  
Several common stereo estimation metrics are used to evaluate the proposed method. End-point-error (\textit{EPE}) is the mean absolute disparity error. \textit{Bad 1} is the percentage of pixels with disparity errors larger than 1 pixel. By converting disparity to depth, we also measure the average absolute depth error (\textit{abs depth err}) and the percentage of depth outliers with absolute error larger than 4mm, which is denoted as \textit{$>$4mm}. To evaluate the performance of our model on objects of different materials, these depth metrics are measured separately on two kinds of objects in the testing dataset using object masks.
Since the RealSense camera outputs a value of zero at areas with high depth uncertainty, metrics are computed in terms of excluding and including uncertain pixels so that the evaluation is in the same completeness level.

\subsection{Comparison with other Methods}
\begin{figure*}[t]
\begin{center}
	\includegraphics*[width=1.0\textwidth]{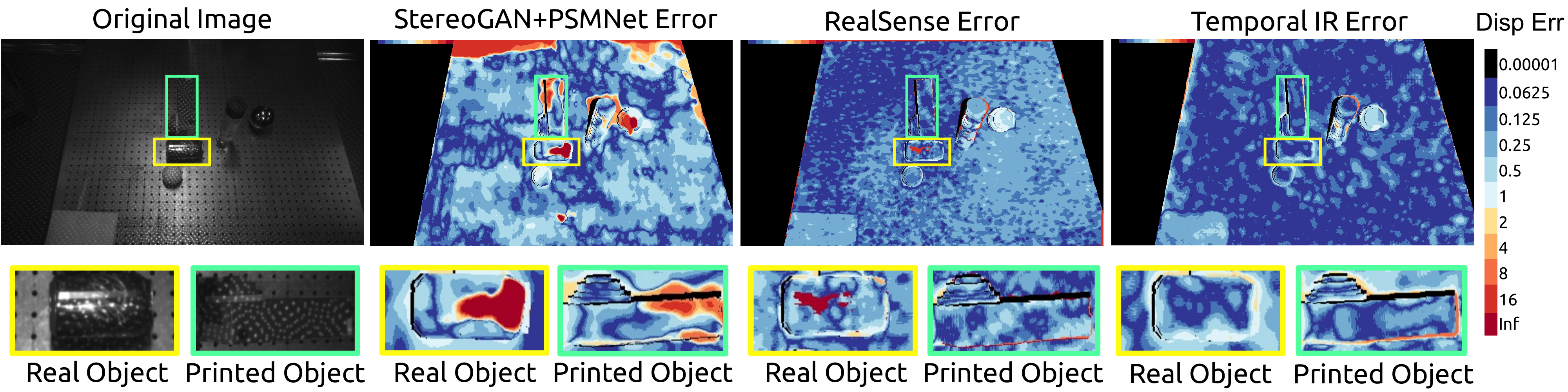}
\end{center}    
  \caption{Comparison of the disparity error map of our method with StereoGAN and RealSense D415. Our method improves disparity accuracy on both 3D-printed objects and real objects.}
\label{fig:errormap}
\end{figure*}

\input{tab/main_res_comparison}

For evaluation, our method is compared with other learning based methods and a decent commercial depth sensor - the RealSense D415. As shown in \cref{tab:main_res}, our method outperforms other methods in all metrics.

\paragraph{Learning-based methods} Our method is best compared with PSMNet~\cite{chang2018pyramid} and StereoGAN~\cite{liu2020stereogan} and we use them as our baselines. To test vanilla PSMNet, we train it on input stereo images with and without active pattern using only the training dataset in the simulation domain and then test it directly in the real testing dataset. As shown in \cref{tab:main_res}, using active pattern can improve the stereo matching accuracy across all metrics  and is beneficial for eliminating the sim-real domain gap. This intuitively makes sense since active light adds pattern to textureless areas which are the most difficult to match.

Furthermore, besides the original StereoGAN\cite{liu2020stereogan}, we extend the StereoGAN architecture by using PSMNet as the disparity prediction backbone, which is denoted as \textbf{StereoGAN+PSMNet}. This improved StereoGAN uses cost volume aggregation in its stereo matching module, which makes it more powerful and comparable with our method. The results show that \textbf{StereoGAN+PSMNet} performs better than \textbf{StereoGAN} in all metrics. Although, when compared with our method, \textbf{StereoGAN+PSMNet} performs considerably worse as the absolute depth error increases from $4.377$mm to $13.762$mm. This is further corroborated by \cref{fig:errormap}, where \textbf{StereoGAN+PSMNet} struggles to predict depth on real objects such as the metal can, which is a specular surface. On the other hand, our \textit{mixed domain learning} method has improved accuracy on these types of objects. This large performance improvement can be attributed to direct supervision in the simulation domain of primitives with random shapes and materials, a well-shaped temporal IR reprojection which accurately locates the correct correspondences, and a more robust pipeline overall since it doesn't use the GAN module.

\paragraph{Intel RealSense D415} To the best of our knowledge, we are the first work to be quantitatively compared with commercial products. The Intel RealSense D415 uses a traditional CENSUS-based stereo matching method \cite{zabih1994non, keselman2017intel}, which has high computation efficiency but will leave uncertain pixels without depth values. Therefore, we report our results on the same completeness levels as RealSense and demonstrate that our method outperforms RealSense in every metric. In \cref{fig:errormap}, RealSense is unable to accurately predict pixels in specular areas, while our method is able to match those pixels well. In addition, for 3D-printed objects, our model also demonstrates lower depth error.

\subsection{Ablation Study}
\label{sec:exp:ablation}
In this section, we validate the effectiveness of each component and design choice  through ablation experiments. 

\paragraph{Reprojection Loss}
\input{tab/ablation_1} 
We compare the network's performance when doing reprojection on different patterns which is shown \cref{tab:ablation1}.

First, we use traditional reprojection loss on input stereo images which simply computes the patch-wise Mean Squared Error (MSE) of the warped images. Second, we use an advanced reprojection loss function from ActiveStereoNet\cite{zhang2018activestereonet}, which uses an LCN module to alleviate the condition where two matched pixels have large residuals due to the distance from the camera and the physical properties of the surface. Third, we experiment with applying reprojection loss on a 2-step IR Pattern. 

For the sake of fairness, we add synthetic ground truth depth supervision to all of the experiments above. The Raw IR reprojection has the worst result because it doesn't take into account the different intensities of IR light of two matched pixels. While LCN IR helps address this issue, it employs reprojection on the continuous local normalized grayscale IR image, which is still affected by environmental illumination and object texture. To tackle this issue, we proposed a reprojection loss on 2-Step IR patterns which shows better performance since the binary pattern eliminates the small residual of two matched pixels. Lastly, since the SNR is low for pixels that are far away from the camera, 2-Step IR cannot properly extract the active light pattern in distant areas. This issue is addressed by our temporal IR patterns. By tracking the intensity difference in the temporal IR image sequence, our approach extracts a more accurate and complete IR pattern. The results prove that our reprojection on temporal IR images is superior to all other reprojection methods.

\paragraph{Simulation Supervision}
In order to investigate the effect of simulation supervision, we implement the experiments listed on \cref{tab:ablation2}. Specifically,  we observe a significant performance drop in the trained model after removing supervision on simulation disparity. Therefore, we can conclude that supervision on simulation domain helps the network achieve better performance.

As mentioned before, the simulation domain can help temporal IR reprojection converge closer and faster to a global minima. Then, temporal IR reprojection serves to further converge to the ground truth disparity. The results in \cref{tab:ablation2} are consistent with the fact that synthetic supervision can further improve the performance. 

\input{tab/ablation_2}

\paragraph{Generalization}
\input{tab/generalization} 
In order to evaluate the generalizability of the learned stereo network trained on the simulated dataset consisting of shape primitives, we construct another simulated dataset using the same objects as in the testing dataset.
As shown in \cref{tab:generalization}, the model trained on the random shape primitives dataset outperforms the model trained on the dataset that contains only shapes and textures that appear in the testing dataset, which validates the claim that greater variation of geometry, texture, and material introduced in our shape primitives dataset leads to superior generalizability of the learned stereo network.

%% file: tab/main_res_comparison.tex
\begin{table*}[th]
\centering
\begin{tabular}{@{}c|c|c|c|c|c|c|c|c@{}}
\toprule
\multicolumn{9}{c}{Excluding uncertain pixels} \\ \hline
\multirow{2}{*}{Method} & EPE (px) $\downarrow$ & Bad 1 $\downarrow$  & \multicolumn{3}{|c}{Abs depth err (mm) $\downarrow$} &\multicolumn{3}{|c}{ $>$ 4mm$\downarrow$} \\ \cline{2-9}
 & All & All & All & Printed & Real& All & Printed& Real\\ \hline
PSMNet \cite{chang2018pyramid} w/o active pattern & 0.664 & 0.187 & 9.218 & 12.600 & 16.467 & 0.478 & 0.686 & 0.836 \\
PSMNet \cite{chang2018pyramid} w/ active pattern & 0.476 & 0.077 & 7.135 & 9.174 & 15.570 & 0.504 & 0.591 & 0.800 \\ \hline
StereoGAN \cite{liu2020stereogan} & 5.603 & 0.741 & 44.284 & 36.892 & 42.105 & 0.925 & 0.915 & 0.931 \\
StereoGAN \cite{liu2020stereogan} + PSMNet \cite{chang2018pyramid} & 2.296 & 0.176 & 13.762 & 22.489 & 37.031 & 0.641 & 0.762 & 0.899\\
RealSense D415 & 0.392 & 0.032 & 5.817 & 7.851 & 15.826 & 0.565 & 0.612 & 0.817 \\ \hline
Ours & \textbf{0.314} & \textbf{0.028} & \textbf{4.377} & \textbf{6.584} & \textbf{14.694} & \textbf{0.335} & \textbf{0.437} & \textbf{0.725} \\
\bottomrule
\multicolumn{9}{c}{Including uncertain pixels} \\ \hline
\multirow{2}{*}{Method} & EPE (px)$\downarrow$ & Bad 1$\downarrow$& \multicolumn{3}{|c}{Abs depth err (mm)$\downarrow$} &\multicolumn{3}{|c}{ $>$ 4mm $\downarrow$} \\ \cline{2-9}
 & All & All & All & Printed& Real& All & Printed& Real\\ \hline
 
PSMNet \cite{chang2018pyramid} w/o active pattern & 0.698 & 0.194 & 9.530 & 12.987 & 16.960 & 0.485 & 0.689 & 0.840 \\
PSMNet \cite{chang2018pyramid} w/ active pattern & 0.513 & 0.084 & 7.444 & 9.580 & 16.745 & 0.510 & 0.595 & 0.804 \\ \hline
StereoGAN \cite{liu2020stereogan} & 5.765 & 0.744 & 44.747 & 36.703 & 42.220 & 0.926 & 0.915 & 0.932\\
StereoGAN \cite{liu2020stereogan} + PSMNet \cite{chang2018pyramid} & 2.472 & 0.185 & 14.318 & 22.818 & 37.753 & 0.645 & 0.764 & 0.902\\
RealSense D415 & 1.793 & 0.056 & 8.159 & 9.891 & 22.492 & 0.576 & 0.621 & 0.835 \\ \hline
Ours & \textbf{0.364} & \textbf{0.035} & \textbf{4.736} & \textbf{7.057} & \textbf{16.191} & \textbf{0.342} & \textbf{0.444} & \textbf{0.740} \\
\bottomrule
\end{tabular}
\caption{
Performance of different state of the art learning-based stereo, commercial depth sensor and our method on the real testing dataset
} 
\label{tab:main_res}
\end{table*}

%% file: tab/ablation_1.tex
\begin{table}[t]
\centering
\begin{tabular}{c|c|c}
\toprule
Pattern & Abs depth err (mm) $\downarrow$ & $>$ 4mm $\downarrow$\\ \hline 
Raw IR & 32.166 & 0.638 \\
LCN IR \cite{zhang2018activestereonet} & 10.598 & 0.512 \\
2-Step IR & 4.697 & 0.373 \\
Temporal IR & \textbf{4.377} & \textbf{0.335} \\
\bottomrule
\end{tabular}
\caption{Comparison of self-supervised reprojection loss on different patterns }
\label{tab:ablation1}
\end{table}

%% file: tab/ablation_2.tex
\begin{table}[t]
\centering
\begin{tabular}{c|c|c}
\toprule
\multicolumn{3}{c}{Without sim ground-truth} \\ \hline
Reprojection & Abs depth err (mm) $\downarrow$ & $>$ 4mm $\downarrow$\\ \hline
Raw IR & 43.326 & 0.716 \\
Temporal IR & \textbf{4.729} & \textbf{0.367} \\ \bottomrule
\multicolumn{3}{c}{With sim ground-truth} \\ \hline
Reprojection &  Abs depth err (mm) $\downarrow$ & $>$ 4mm $\downarrow$\\ \hline
Raw IR & 32.166 & 0.638 \\
Temporal IR & \textbf{4.377} & \textbf{0.335} \\
\bottomrule
\end{tabular}
\caption{Comparison of disparity supervision in the simulation domain with different self-supervised reprojection loss}
\label{tab:ablation2}
\end{table}

%% file: tab/generalization.tex
\begin{table}[t]
\centering
\begin{tabular}{c|c|c}
\toprule
Simulation Dataset &  Abs depth err (mm) $\downarrow$ & $>$ 4mm $\downarrow$\\ \hline 
Testing objects & 4.388 & 0.347 \\
Shape primitives & \textbf{4.377} & \textbf{0.335} \\
\bottomrule
\end{tabular}
\caption{Performance of network trained on different simulation datasets, `Testing objects' consists of only objects in the testing dataset, `Shape primitives' consists of shape primitives of different size, texture and material}
\label{tab:generalization}
\end{table}

%% file: sec/Conclusion.tex
In this paper, we propose a novel end-to-end training framework, \textit{mixed domain learning}, for learning-based active stereo that surpasses commercial depth sensors and state-of-the-art methods in the real world without any real depth annotation. 
One limitation of our work is that we only evaluate its effectiveness on one type of active stereovision sensor. Further study is needed to understand the extent to which our learned stereo network transfers to other out-of-distribution real datasets and types of sensors. Additionally, in order for this framework to be useable in real applications, we would need to investigate how to accelerate network inference to achieve real-time depth predictions.